# BI-RADS-NET: AN EXPLAINABLE MULTITASK LEARNING APPROACH FOR CANCER DIAGNOSIS IN BREAST ULTRASOUND IMAGES

*Boyu Zhang[1], Aleksandar Vakanski[2],* Member, IEEE, *Min Xian[3],* Member, IEEE

[1]Institute for Modeling Collaboration and Innovation, University of Idaho, Moscow, USA
[2]Department of Nuclear Engineering and Industrial Management, University of Idaho, Idaho Falls, USA
[3]Department of Computer Science, University of Idaho, Idaho Falls, USA
{boyuz, vakanski, mxian}@uidaho.edu

## ABSTRACT

In healthcare, it is essential to explain the decision-making process of machine learning models to establish the trustworthiness of clinicians. This paper introduces BI-RADS-Net, a novel explainable deep learning approach for cancer detection in breast ultrasound images. The proposed approach incorporates tasks for explaining and classifying breast tumors, by learning feature representations relevant to clinical diagnosis. Explanations of the predictions (benign or malignant) are provided in terms of morphological features that are used by clinicians for diagnosis and reporting in medical practice. The employed features include the BI-RADS descriptors of shape, orientation, margin, echo pattern, and posterior features. Additionally, our approach predicts the likelihood of malignancy of the findings, which relates to the BI-RADS assessment category reported by clinicians. Experimental validation on a dataset consisting of 1,192 images indicates improved model accuracy, supported by explanations in clinical terms using the BI-RADS lexicon.

## 1. INTRODUCTION

Explaining the behavior of machine learning (ML) models increases the trustworthiness and confidence in the predictions [1, 2]. The importance of ML explainability in healthcare cannot be overemphasized, because clinicians require to know the 'reason' behind the prediction to inform diagnosis, risk assessment, treatment planning, etc. [3, 4]. In modern computer-aided diagnosis (CAD) systems, it is preferred to adopt ML algorithms that provide explanations of models' information processing aligned with the medical diagnosis process [5]. However, current CAD systems for cancer diagnosis typically output the category of identified tumors (benign or malignant) and/or their location in the image (or the mask of the tumor overlaid over the background tissues). I.e., CAD systems often lack means for associating the outputs of algorithms with the underlying descriptors used by clinicians for image interpretation and diagnosis.

In this work, we introduce a novel approach for explainable breast cancer diagnosis based on the Breast Imaging – Reporting and Data System (BI-RADS) lexicon for breast ultrasound (BUS) [6]. The BI-RADS lexicon standardizes clinical interpretation and reporting, by using a set of descriptors (orientation, shape, margin, echo-pattern, and posterior features) and assessment categories (ranging from 0 to 6, designating increasing likelihood of malignancy).

We propose BI-RADS-Net, a deep learning network comprising a series of convolutional layers for feature extraction, followed by fully-connected layers for output prediction. The architecture contains multiple classification branches that output five BI-RADS descriptors and the tumor class (benign or malignant), and a regression branch that outputs the likelihood of malignancy. The choice of a regression branch in the architectural design was motivated by the widely-reported high inter-observer variability in assigning the BI-RADS assessment categories (in particular, the poor reproducibility for the subcategories 4A, 4B, and 4C) [7, 8]. The validation results on a dataset of 1,192 BUS images indicate that the proposed multitask approach improves the performance in comparison to a single-task approach. In addition, the parameters of the feature extraction layers are shared between all branches in the network, which allows explaining the feature maps that are used for tumor classification in terms of the respective BI-RADS descriptors and the likelihood of malignancy.

Prior work in the literature has designed neural network (NN) architectures for predicting the BI-RADS category of tumors in BUS images, however, without addressing the

Research reported in this publication was supported by the National Institute of General Medical Sciences of the National Institutes of Health under Award Number P20GM104420. The content is solely the responsibility of the authors and does not necessarily represent the official views of the National Institutes of Health.

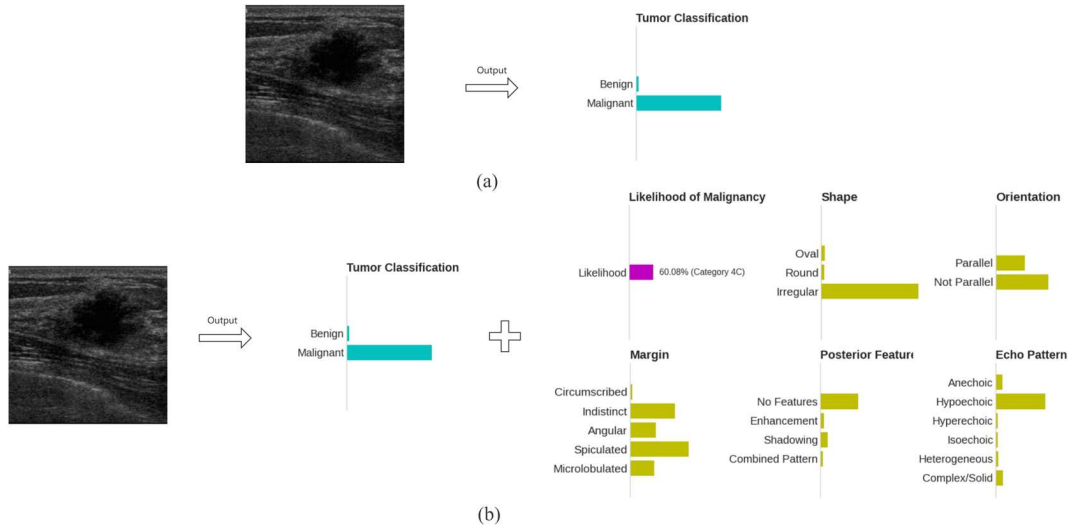

**Fig. 1.** (a) Typical output of a conventional BUS CAD system; (b) Output of the proposed explainable BUS CAD system for the same two images. The bars in the sub-figures indicate the predicted class probabilities by the CAD systems.

model explainability [9, 10]. Also, several authors employed the BI-RADS descriptors for explaining NN models for breast cancer diagnosis in mammography images [11–13]. To the best of our knowledge, the paper by Zhang *et al*. [14] is the only work that used the BI-RADS lexicon for explainability of NN models for BUS. Although our work has similarities to the approach in [14]—both rely on multitask learning framework and the BI-RADS terminology—there are also multiple differences between the two. Specifically, unlike [14], besides the tumor class, our approach outputs all five BI-RADS descriptors and the likelihood of malignancy (see Fig. 1) which are explicitly associated with the clinical features used for BUS interpretation.

The main contributions of our approach include:

• An explainable multitask learning approach that concurrently outputs the BI-RADS descriptors, BI-RADS likelihood of malignancy, and the tumor class (Fig. 1);

• A network architecture with a regression branch to handle the inherent noise in the ground-truth labels for the BI-RADS categories, caused by inter-observer variability;

• Increased tumor classification accuracy, via learning feature representations related to clinical descriptors; and

• The capacity to assess uncertainties in the model outputs for individual BUS images, based on (dis)agreement in the predictions by the different branches of the model.

## 2. RELATED WORK

### 2.1. Explainable ML for medical image analysis

The majority of related work on explainable ML in CAD employed model saliency as a means for post-hoc visual explainability, utilized to outline important regions in images that contributed the most to the model prediction [15, 16]. Similarly, the attention mechanism in NNs has been used for segmentation of organs and lesions [17]. TIRADS clinical features were also leveraged for explainable ML of thyroid nodules diagnosis [18]. Furthermore, existing models focused on concurrently processing medical images and creating textual reports similar to clinicians' reports when interpreting medical images [19, 20]. In general, explainable ML in healthcare introduces unique challenges that reverberate with the very challenges in medical image analysis. These include small datasets, low contrast, complex image formats (e.g., 3D or 4D image modalities), large image size and high resolution, and important details removed by preprocessing techniques. In addition, the level of risk and responsibility for explainable ML in healthcare are uniquely distinct, since the decisions may affect the lives of patients.

### 2.2. Explainable ML for breast cancer diagnosis

A body of work investigated the problem of explainable ML for breast cancer CAD. Shen *et al*. [21] introduced an explainable ML classifier that indicated the location of suspected lesions in mammograms. Similarly, Wu *et al*. [11] proposed DeepMiner, an NN architecture for outputting both the tumor class and text explanations using the BI-RADS lexicon for mammography. Kim *et al*. [12, 13] proposed NN models that employed the shape and margin of tumors in mammograms for predicting the class label and BI-RADS category. A key shortcoming of these approaches in mammography is using only two or three BI-RADS descriptors, which often lack sufficient information to fully explain the intricate process of tumor diagnosis.

Similarly, despite the impressive recent progress in BUS tumor classification and segmentation, the explainability for BUS CAD has been less extensively explored by the research community. The only approach on explainability for BUS CAD was proposed by Zhang *et al*. [14]. The authors introduced a preprocessing step for emphasizing the BI-RADS descriptors of shape and margin in BUS images, and an encoder-decoder NN was used for predicting the tumor class and reconstructing the input image. A limitation of the approach in [14] is that only the shape and margin were used for tumor classification, and the class probabilities of these two descriptors were not output by the model (to explain the tumor classification). Approaches that concentrated on generating textual reports for explaining NN models for BUS [22], as well as for identifying explainable salient regions in breast histopathology images [23] were also proposed in the literature. Despite these latest efforts, explainability of CAD systems for breast cancer diagnosis is still an open research problem that requires further attention.

## 3. PROPOSED METHOD

### 3.1. BI-RADS lexicon

BI-RADS is a risk assessment system introduced by the American College of Radiology to standardize the assessment, reporting, and teaching of breast imaging. It applies to mammography, ultrasound, and MRI. The BI-RADS lexicon assigns a mass finding to one of seven assessment categories shown in Table I, that designate a likelihood of malignancy in the 0-100% range. E.g., BI-RADS category 0 is assigned to cases with incomplete imaging, whereas BI-RADS category 6 is assigned to biopsy-validated malignant cases. For BI-RADS category 4, three sub-categories were introduced that designate low suspicion (4A), moderate suspicion (4B), and high suspicion of malignancy (4C). The BI-RADS categories are important for cancer risk management, where BI-RADS 3 patients are scheduled for short-term follow-up imaging, whereas BI-RADS 4 and 5 patients undergo diagnostic biopsy.

Besides the assessment categories, the BI-RADS lexicon provides terminology to describe different features of the mass findings in BUS. The BI-RADS descriptors for BUS are shown in Table II, and include shape, orientation, margin, echo pattern, and posterior features. The table also lists the standardized terms used for the classes of the descriptors.

### 3.2. Data

The presented approach is evaluated with 1,192 BUS images, obtained by combining two datasets, referred to as BUSIS [24] and BUSI [25]. The BUSIS dataset consists of 562 images, of which 306 images contain benign and 256 contain malignant tumors. From the BUSI dataset we used a subset of 630 images containing tumors, of which 421 have benign and 209 have malignant tumors. The combined dataset has class imbalance, as it consists of 727 benign and 465 malignant images. All images were annotated with ground-truth labels for the tumor class, BI-RADS descriptors, and BI-RADS assessment category. Image acquisition for the two datasets was performed by different types of imaging ultrasound devices and with different populations of patients. Although this reduces the classification performance of the DL models, on the other hand, it improves the robustness of the approach to variations in unobserved images. The details regarding the BUSIS and BUSI datasets are provided in the respective publications [24] and [25].

TABLE I. BI-RADS ASSESSMENT CATEGORIES

| Category | Assessment | Likelihood of Malignancy | Management |
|---|---|---|---|
| 0 | Incomplete | NA | Additional imaging required |
| 1 | Negative | No cancer detected | Annual screening |
| 2 | Benign | 0% | Annual screening |
| 3 | Probably benign | 0-2% | Follow-up in 6 months |
| 4A | Suspicious | 2-10% | Tissue diagnosis |
| 4B | Suspicious | 10-50% | Tissue diagnosis |
| 4C | Suspicious | 50-95% | Tissue diagnosis |
| 5 | Malignant | >95% | Tissue diagnosis |
| 6 | Biopsy-proven malignancy | Cancer present | Surgical excision |

TABLE II. BI-RADS DESCRIPTORS FOR BUS IMAGES

| BI-RADS Descriptors | Descriptors Class |
|---|---|
| Shape | Oval, Round, Irregular |
| Orientation | Parallel, Not parallel |
| Margin | Circumscribed, Not circumscribed (Indistinct, Angular, Microlobulated, Spiculated) |
| Echo pattern | Anechoic, Hypoechoic, Isoechoic, Hyperechoic, Complex cystic and solid, Heterogeneous |
| Posterior features | No posterior features, Enhancement, Shadowing, Combined pattern |

### 3.3. Network architecture

The architecture of BI-RADS-Net is depicted in Fig. 2, and it consists of two major components: a shared backbone network and task-specific networks entailing branches for predicting the BI-RADS descriptors, BI-RADS likelihood of malignancy, and the tumor category. The backbone network employs convolutional and max-polling layers for extracting relevant features in input BUS images. The learned feature maps are employed by the BI-RADS descriptors branch to predict the five descriptors from Table II. The outputs for the BI-RADS descriptors are concatenated with the feature maps from the base network and are fed to a regression branch to predict the likelihood of malignancy. The regression branch outputs a continuous value ranging from 0% to 100%. The tumor classification branch merges the features maps from the backbone network and the other two branches to output a binary benign or malignant class label.

The ground-truth labels for the BI-RADS descriptors are as listed in Table II. I.e., shape has 2 classes (parallel and not parallel), orientation has 3 classes, echo pattern has 6 classes, and posterior features has 4 classes. The margin descriptor can have multiple annotations. For instance, the margin in Fig.

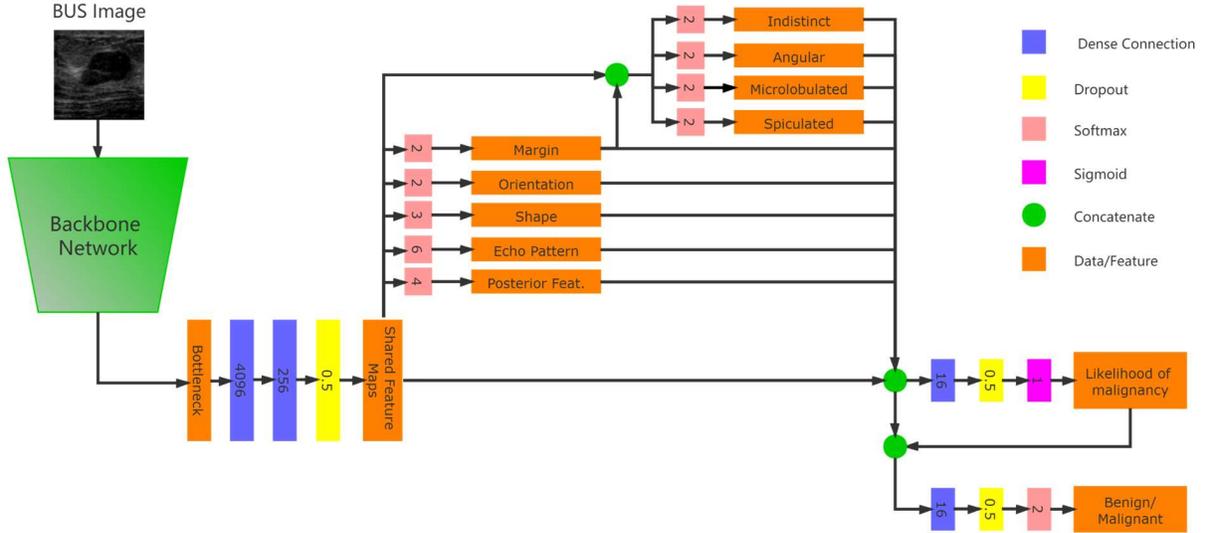

**Fig. 2.** Network architecture of the proposed BI-RADS-Net for BUS CAD.

1(b) is not circumscribed, and it is both indistinct and spiculated. Therefore, the first branch for the margin in BI-RADS-Net has only 2 classes (circumscribed and not circumscribed), and afterward, another sub-branch is introduced that outputs binary values for the indistinct, angular, microlobulated, and spiculated margin sub-classes.

For the likelihood of malignancy branch, as ground-truth we used continuous values corresponding to the BI-RADS assessment categories shown in Table 1. We adopted the median likelihood of malignancy, as follows: Category 3 – 1%, Category 4A – 6%, Category 4B – 30%, Category 4C – 72.5%, and Category 5 – 97.5%. Predicting continuous values for the likelihood of malignancy using a regression branch instead of categorical variables enables the network to deal with inter-observer variability in the BI-RADS category labels. Note also that the BUSIS and BUSI datasets do not contain images with BI-RADS 0, 1, 2, or 6 categories.

In the multitask model, Task 1 to 5 are the BI-RADS descriptors, Task 6 to 9 are the sub-classes for the margin BI-RADS descriptor, Task 10 is the BI-RADS likelihood of malignancy, and Task 11 is the tumor classification branch. For each task $k$, the network loss function is denoted by $\mathcal{L}_k(X_k, Y_k)$, where $X_k$ is the predicted value and $Y_k$ is the ground-truth label (for classification) or value (for regression). Since the outputs of the likelihood of malignancy branch (Task 10) and the tumor classification branch (Task 11) both reflect the level of risk that the present tumor in the image is malignant, we added an additional loss term $\mathcal{L}_a$ to enforce an agreement between the two branches. The total loss of the model is calculated as the weighted sum of all tasks, that is, $\mathcal{L} = \sum_{i=1}^{K} \lambda_i \mathcal{L}_i(X_i, Y_i) + \lambda_a \mathcal{L}_a(|X_{11}-X_{10}|, |Y_{11}-Y_{10}|)$. The symbol $\lambda_i$ denotes the weight coefficient of task $i$, $K = 11$ is the number of tasks, and $\lambda_a$ is the weight coefficient for the $\mathcal{L}_a$ term. Cross-entropy loss and mean-square error loss are used for the classification and regression branches, respectively.

### 3.4. Implementation details

The size of input images to the network was set to 256×256 pixels. In order to prevent distortion of the morphological features, such as shape and orientation, the original BUS images were first cropped to the largest squared segment that encompasses the tumor, and afterward, the cropped segment was resized to 256×256 pixels. If the original images were directly resized to 256×256 pixels, the labels for the shape and orientation for some images would be incorrect (e.g., the shape of some tumors can change from oval to round when wide rectangular images are resized to square images).

Next, for each BUS image comprising a single gray channel, we added two additional channels. One channel was obtained by performing histogram equalization to the gray channel, and another channel was obtained by applying smoothing to the gray channel. We found that this simple preprocessing step was beneficial to improving the model performance. One possible explanation is that histogram equalization and smoothing reduced the variations across the images in BUSIS and BUSI datasets, and resulted in a more uniformly distributed set of images.

We used five-fold cross-validation, i.e., the images were split into 80% training and 20% testing sets. Further, 15% of the images in the training set were used for validation.

For the backbone network we used the encoder of a VGG-16 model, initialized with parameters pretrained on the ImageNet dataset. The parameters in all network layers were updated during training. We applied various types of data augmentation techniques, including zoom (20%), width shift (10%), rotation (5 degrees), shear (20%), and horizontal flip. We used a batch size of 6 images. The models were trained by using adaptive moment estimator optimized (Adam), with an initial learning rate of $10^{-5}$, which was reduced to $10^{-6}$ if the loss of the validation set did not reduce for 15 epochs. The training was stopped when the loss of the validation set did not reduce for 30 epochs. For the loss weight coefficients $\lambda_1$ to $\lambda_{11}$, we adopted the following values: (0.2, 0.2, 0.2, 0.2, 0.2, 0.1, 0.1, 0.1, 0.1, 0.2, 0.5). That is, the largest weight was assigned to the tumor class branch. The weight $\lambda_a$ for the loss term $\mathcal{L}_a$ was set to 0.2 as well.

TABLE III. ABLATION STUDY, REGARDING THE IMPACT OF DIFFERENT COMPONENTS IN THE NETWORK DESIGN ON THE PERFORMANCE

| Method | Tumor Class | | | BI-RADS Descriptors | | | | | Likelihood of Malignancy | |
|---|---|---|---|---|---|---|---|---|---|---|
| | Accuracy | Sensitivity | Specificity | Shape | Orientation | Margin | Echo Pat. | Post. Feat. | $R^2$ | MSE |
| **BI-RADS-Net** | **0.889** | 0.838 | **0.923** | 0.816 | **0.872** | **0.873** | **0.825** | 0.830 | **0.671** | **0.153** |
| Without Augmentation | 0.868 | 0.789 | 0.919 | **0.832** | 0.848 | 0.855 | 0.804 | 0.828 | 0.648 | 0.159 |
| Without Pretraining* | 0.828 | 0.746 | 0.881 | 0.773 | 0.804 | 0.794 | 0.726 | 0.731 | 0.592 | 0.171 |
| Single Channel Images* | 0.817 | 0.726 | 0.875 | 0.764 | 0.809 | 0.792 | 0.720 | 0.739 | 0.582 | 0.173 |
| Without Image Cropping* | 0.799 | 0.711 | 0.855 | 0.755 | 0.788 | 0.774 | 0.716 | 0.729 | 0.528 | 0.184 |
| ResNet Backbone | 0.883 | **0.841** | 0.909 | 0.816 | 0.850 | 0.868 | 0.813 | **0.831** | 0.664 | 0.155 |
| EfficientNet Backbone | 0.856 | 0.826 | 0.904 | 0.819 | 0.858 | 0.847 | 0.795 | 0.826 | 0.667 | 0.154 |

\* The ablation steps are progressively applied, i.e., the model without augmentation is afterward evaluated without pretrained weights, etc.

TABLE IV. MULTITASK APPROACH EVALUATION

| Method | Tumor Class | | |
|---|---|---|---|
| | Accuracy | Sensitivity | Specificity |
| Single Branch Tumor Class | 0.864 | 0.795 | 0.908 |
| + Margin | 0.878 | 0.817 | 0.918 |
| + Orientation + Shape | 0.883 | 0.821 | 0.922 |
| + Echo pattern + Post. feat. | 0.887 | 0.831 | 0.923 |
| + Likelihood of Malignancy = **BI-RADS-Net** | **0.889** | 0.838 | 0.923 |

**4. EXPERIMENTAL RESULTS**

The results of an ablation study performed to evaluate the impact of the different components in the design of BI-RADS-Net are shown in Table III. The ablation study assesses the contributions by data augmentation, pretrained network parameters on the ImageNet dataset, additional image channels with histogram equalization and smoothing, and cropping the original images to square-size segments. The results indicate that the network achieved over 80% accuracy for all five BI-RADS descriptors, whereas the tumor class accuracy reached 88.9%. Due to space limitation, the results for the margin sub-classes are not presented (for all 4 sub-classes the accuracy overpassed 80%). Table III also presents a comparison for the presented model with a VGG backbone to ResNet50 and EfficientNet-B0 backbones.

Table IV presents the evaluation of the effectiveness of the multitask learning approach. The accuracy of a single-task model for tumor classification is 86.4%, and it increases to 88.9% for the model with multiple branches. Thus, the information provided by the BI-RADS descriptors benefits the tumor classification branch. In general, the largest positive correlation with the BUS tumor class is reported in the literature for the margin descriptor, followed by shape and orientation. Echo pattern and posterior features have lower correlations comparatively; however, the two descriptors are still important for BUS interpretation and diagnosis. The contribution by the likelihood of malignancy branch to the tumor class prediction in Table IV is lower compared to the other branches. Examples of outputs generated by BI-RADS-Net are shown in Fig. 1.

The objective of our approach is to provide explanations for the classification of BUS images containing tumors into a benign or malignant class. Explainability is achieved by reporting the BI-RADS descriptors and likelihood of malignancy. We hold that this information would be beneficial and valuable to clinicians for interpreting BUS images. First, this information provides a link between the information processing by the CAD model and medical diagnosis by clinicians. Namely, clinical interpretation involves observing the shape, orientation, margin, echo pattern, and posterior features of masses, in combination with associated features (duct, skin changes), special cases (implants), and considering additional information, such as the patient medical history, age, lifestyle, or known risk factors. Second, the provided information can be helpful for the reporting phase. Third, evaluating the uncertainties in the ML predictions on individual BUS images is especially challenging: whenever there is a discrepancy between a clinician's interpretation and the CAD tumor class prediction on an individual BUS image, the clinician might be suspicious about the CAD prediction. Providing explanations via the BI-RADS descriptors and the BI-RADS likelihood of malignancy can assist clinicians in understanding the level of uncertainties in the model's output on individual BUS images. Subsequently, the provision of explainability using the BI-RADS lexicon can increase the trustworthiness of clinicians in the CAD systems.

The proposed approach differs from the common post-hoc explainability approaches, where explanations of the decision-making for a model are provided after the training phase is completed. Instead, we use a single end-to-end deep learning model that furnishes explainability concurrently with the training/testing phases. We justify such an approach because we relied on a clinically validated set of visual features—BI-RADS descriptors—to explain BUS analysis.

We posit that explainability is task-dependent and audience-dependent, and therefore, requires ML models designed for specific tasks and targeted to end-users. For instance, the practical relevance of our proposed explainable model for BUS would diminish for other tasks, because they employ different image features for representation learning. Likewise, our approach may not provide adequate explainability to a data scientist without medical knowledge, or to patients. In this respect, our model is designed for providing explanations to and assisting BUS clinicians.

And, on a separate note, although it is possible to train individual single-task NNs for each BI-RADS descriptor to provide explainability, sharing the encoder by all branches in BI-RADS-Net ensures that the features maps used for tumor classification are relevant to the BI-RADS descriptors and likelihood of malignancy.

## 5. CONCLUSION

This paper describes BI-RADS-Net, a multitask deep learning model for explainable BUS CAD using the BI-RADS lexicon. The network architecture consists of multiple classification and regression branches that output the tumor class, five BI-RADS descriptors, and the likelihood of malignancy (in relation to the BI-RADS assessment category). The prediction of the tumor class (benign or malignant) made by the model is presented in a form that is understandable to clinicians via the BI-RADS descriptors of mass findings and the risk of malignancy. The proposed approach departs from the traditional post-hoc techniques for explainable deep learning, and instead integrates the explainability directly into the outputs generated by the model. The reason such an explainable approach can be effective is because we rely on a predefined set of morphological mass features, adopted from the BI-RADS lexicon. Furthermore, such an approach is aligned with the visual perception and reasoning process by clinicians when interpreting BUS images. Conclusively, the proposed approach is designed to assist clinicians in interpretation, analysis, and reporting in BUS. In future work, we will conduct a study for qualitative assessment of the level of explainability of our approach with BUS clinicians via structured interviews and questionnaires.